\documentclass[letterpaper, 10 pt, conference]{ieeeconf}  

\IEEEoverridecommandlockouts                              

\usepackage{graphicx}

\usepackage{amsmath} 
\usepackage{amssymb}  
\usepackage{amsfonts}
\usepackage{mathrsfs}
\usepackage{wrapfig}
\usepackage{cite}
\usepackage{graphicx}
\usepackage{textcomp}
\usepackage{xcolor}
\usepackage[capitalise]{cleveref}
\usepackage[list=on,listformat=simple]{subcaption}
\usepackage{preambles}
\usepackage{algorithm} 
\usepackage{algpseudocode}

\title{\LARGE \bf
Template Model Inspired Task Space Learning for Robust Bipedal Locomotion}
\author{Guillermo A. Castillo$^{1}$,  Bowen Weng$^{1}$, Shunpeng Yang$^{2}$, Wei Zhang$^{2}$, and Ayonga Hereid$^{3}$
\thanks{This work was supported in part by the National Science Foundation
under grant FRR-21441568, and by the National Natural Science Foundation of China under Grant No. 62073159.}%
\thanks{$^{1}$Electrical and Computer Engineering, Ohio State University, Columbus, OH, USA;  {\tt\footnotesize \{castillomartinez.2,weng.172\}@osu.edu.}}
\thanks{$^{2}$SUSTech Institute of Robotics, Southern University of Science and Technology (SUSTech), China; {\tt\footnotesize 11930364@mail.sustech.edu.cn, zhangw3@sustech.edu.cn.}}
\thanks{$^{3}$Mechanical and Aerospace Engineering, Ohio State University, Columbus, OH, USA. {\tt\footnotesize hereid.1@osu.edu.}}%
}

\begin{document}

\maketitle


\begin{abstract}

This work presents a hierarchical framework for bipedal locomotion that combines a Reinforcement Learning (RL)-based high-level (HL) planner policy for the online generation of task space commands with a model-based low-level (LL) controller to track the desired task space trajectories.
Different from traditional end-to-end learning approaches, our HL policy takes insights from the angular momentum-based linear inverted pendulum (ALIP) to carefully design the observation and action spaces of the Markov Decision Process (MDP). This simple yet effective design creates an insightful mapping between a low-dimensional state that effectively captures the complex dynamics of bipedal locomotion and a set of task space outputs that shape the walking gait of the robot. The HL policy is agnostic to the task space LL controller, which increases the flexibility of the design and generalization of the framework to other bipedal robots.
This hierarchical design results in a learning-based framework with improved performance, data efficiency, and robustness compared with the ALIP model-based approach and state-of-the-art learning-based frameworks for bipedal locomotion.
The proposed hierarchical controller is tested in three different robots, Rabbit, a five-link underactuated planar biped; Walker2D, a seven-link fully-actuated planar biped; and Digit, a 3D humanoid robot with 20 actuated joints. The trained policy naturally learns human-like locomotion behaviors and is able to effectively track a wide range of walking speeds while preserving the robustness and stability of the walking gait even under adversarial conditions. 

\end{abstract}


\section{Introduction}\label{sec:intro}

Robust bipedal robot locomotion presents a challenging problem for robotics research due to the complexity of high dimensional models, unilateral ground contacts, and nonlinear and hybrid dynamics. Common methods applied in bipedal locomotion rely on solving optimization problems using the robot's full-order or reduced-order model to find feasible trajectories that realize stable walking gaits. In general, using full-order models results in computationally expensive problems that cannot be solved in real-time~\cite{hereid2018dynamic, grizzle2014models}. 
To reduce the computation time, reduced-order models are used to capture the dynamics of the full-order system and plan trajectories for the robot's center of mass (CoM) and end-effectors. 
However, the assumptions made on reduced-order models such as a constant CoM height limit their performance on dynamic locomotion behaviors and their accuracy to predict the behavior of the real robot under certain conditions. Recently, the angular momentum-based linear inverted pendulum (ALIP) has been presented as an improved alternative to the Linear Inverted Pendulum (LIP) to predict the evolution of the model's state, demonstrating in simulation and hardware experiments that the angular momentum about the contact point can be more accurately predicted than the CoM velocity~\cite{gong2021onestep, gibson2021terrain}.


With the recent success of deep learning in tackling challenging control problems, machine learning-based approaches have exploited advances in physics simulators and computing power to learn locomotion policies through more structured learning frameworks. Learning from motion references has become a popular choice to exploit large amounts of data to train walking policies. The data is obtained from motion capture systems, public motion data sets, or even from video clips, and it is used as goal references in the reward design~\cite{peng2017deeploco, peng2018deepmimic, starke2022deepphase}. 

Other learning methods rely on using optimization to obtain a single feasible reference trajectory \cite{xie2018feedback, siekmann2020learning}, or libraries of reference trajectories ~\cite{li2021reinforcement, green2021learning} to guide the learning. However, these approaches require large amounts of data, and the learned policy often lacks interpretability and control over the parameters of the walking gait. This makes it difficult to adjust the policy during the sim-to-real transition. 
As an alternative, more complex frameworks have been proposed to combine learning algorithms with model-based controllers. The authors in~\cite{castillo2019reinforcement} take insights from the Hybrid Zero Dynamics (HZD) to learn joint trajectories for planar robots. In \cite{jimenez2022neural}, an HZD-based approach is used to learn a policy that satisfies Control Barrier Functions (CBF) defined on the reduced-order dynamics. 

In this work, we propose a hierarchical RL-based approach to address bipedal locomotion in underactuated and fully actuated robots. At the HL stage, RL is used to train policy that learns task space commands for different walking speeds. At the LL, a model-based nonlinear controller is implemented to track the trajectories generated by the HL planning. Several RL-based approaches have been already proposed to exploit hierarchical structures. In~\cite{duan2021learning, green2021learning}, a task space policy is trained to walk at different speeds. However, the method relies on solving a series of optimization problems using the Spring Linear Inverted Pendulum to create a gait library that is used as a reference for the reward and the target end-effector positions. The policy learns residual terms that are added to the task space references~\cite{duan2021learning} or joint space references~\cite{green2021learning}. Different from these approaches, our method directly learns a set of task space actions that completely characterize the dynamic walking gait without the need for previously computed reference trajectories. Moreover, we use different state and action spaces that significantly simplify the complexity of the learning problem and can be generalized to both unactuated and underactuated robots. 

In our previous work~\cite{castillo2022reinforcement}, a cascade structure is implemented to compensate the learned trajectories with feedback regulators to increase the robustness of the walking gait ~\cite{castillo2020hybrid, castillo2021robust}.
Although the method was successfully tested in hardware, the interpretability of the learned policy was limited by the complex structure imposed over the input-output mapping of the RL policy and the addition of compensation terms on top of the learned joint trajectories. On the one hand, the integration of the feedback regulators improves the robustness and sim-to-real transfer of the learned policy. On the other hand, it makes it difficult to identify the actual contribution of the learned policy to the robustness of the walking gait. This results in a policy limited to naturally exploiting the state and action spaces and a restricted walking speed range with the robot Digit, e.g., $v_x \in [-0.5, 0.5]$ m/s.

In this work, we propose a more efficient and clean framework that completely decouples the HL learning policy from the LL controller with better insights into the selection of the state and action spaces that results in improved sample efficiency and interpretability of the policy. We demonstrate the proposed framework is general for 2D and 3D bipedal robots and can be applied even in the case of underactuated robots. Moreover, we show that the learned policy achieves enhanced performance and robustness compared with our previous work ~\cite{castillo2022reinforcement}. 

    The main contributions of this paper are as follows:
    1) \textbf{A simple, efficient, and general} hierarchical learning framework that fully decouples the HL planner from the LL feedback controller. Different from other task space learning approaches, our method \textbf{(i)} uses a reduced-order state for the RL, \textbf{(ii)} learns to walk from scratch, and \textbf{(iii)} computes a set of task space actions that fully characterize dynamic walking gaits. The selection of inputs and outputs is general to bipedal robots of different morphology and degrees of freedom. We show results for actuated and underactuated 2D (Rabbit, Walker2D) and 3D robots (Digit). 
 %
    
    2) \textbf{Insightful design of the RL state space}. We use the ALIP state and speed tracking information to design a reduced-order state space for the RL that captures the complex dynamics of bipedal locomotion while simplifying the learning process.  

    
    
    3) \textbf{Enhanced flexibility of the policy} to naturally exploit the nonlinear dynamics of bipedal locomotion. By including the desired step length, torso orientation, and CoM's height in the action space of the RL, the policy is not restricted to particular behaviors. This allows the policy to learn natural behaviors seen in dynamic locomotion without enforcing them during training.
    
    4) \textbf{A robust locomotion controller} that accurately tracks a wide range of walking speeds, even under external disturbances and challenging terrains, with inclinations up to 20 degrees for both underactuated and fully-actuated robots.

\section{Preliminaries and Problem Formulation}\label{sec:problem_formulation}

\subsection{Bipedal locomotion as a hierarchical problem}
In general, the bipedal locomotion problem can be characterized as a hybrid system determined by a collection of phases of continuous dynamics with discrete events between the transitions of the continuous phases. Formally, the hybrid system model for biped locomotion can be defined as 
\begin{align}\label{eq:hybrid_model}
  \Sigma: \left\{
\begin{array}{lcl}
    \dot{x}  = f(x) + g(x)u + \mathbf{\omega}(x,u) & \hspace{.01cm} & x \in \mathcal{X} \setminus \mathcal{H}\\
    x^+  = \Delta(x^-)       & \hspace{.01cm} & x^- \in \mathcal{H},
  \end{array}
\right.
\end{align}   
where $x \in \mathcal{X} \subseteq \R^n $denotes the robot states, $u \in \mathcal{U} \subseteq \R^m$ is a vector of actuator inputs.   and $\mathbf{\omega} \in \Omega \subseteq \R^w$ a vector of disturbances and uncertainties. 
The switching surface $\mathcal{H}$ is typically the hyper-surface of points corresponding to the height of the swing leg above the ground being zero, and the reset map $\Delta: \mathcal{H} \to \mathcal{X}$ denotes the post-impact state values $x^+$ immediately after switching as a function of the pre-impact state values $x^-$ right before switching.

\begin{figure}[h]
    \begin{center} 
    \includegraphics[width=1\linewidth]{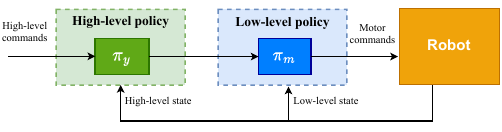}
    \end{center} 
    \caption{Hierarchical structure for bipedal locomotion} 
    \label{fig:hierarchical_structure}
\end{figure}

The control of the bipedal locomotion system described by equation \eqref{eq:hybrid_model} can be formulated as a hierarchical control problem composed by a HL planner and a LL tracking controller. This cascade structure is presented in \figref{fig:hierarchical_structure}. The high level policy ${\pi_y}$
generates trajectories to realize walking gaits according to design parameters and HL commands, e.g., average walking speed, robustness, terrain slope, etc. The LL policy $\pi_m$
computes the actuator inputs to track the desired trajectories commanded by the HL planner. 

The general structure presented in \figref{fig:hierarchical_structure} can characterize most controller formulations used in both model-based and model-free state-of-the-art methods for bipedal locomotion. 
Once the HL trajectories have been generated by the policy, classic model-based control approaches such as feedback linearization, inverse dynamics QP, operational task space controllers, or simple PD controllers can be used to track the desired HL commands. The choice of the LL control policy $\pi_m$ will mostly depend on the action space of the HL policy, e.g., joint space versus task space.




\subsection{Reduced order models for HL planning}
Reduced order models have become a powerful tool for the design of HL planners for bipedal locomotion since they allow using simple dynamical models to characterize biped walking behaviors. 
Recently, the ALIP model has gained attention because of its advantages over LIP to predict the evolution of the state space. The learning-based approach proposed in this work is heavily inspired by recent results using ALIP as a step planner~\cite{gong2021onestep, gibson2021terrain}.



\newsec{Angular Momentum-based Linear Inverted Pendulum (ALIP):}
Considering the states $\{x,L^y\}$, where $x$ is the CoM position in $x$ direction and $L^y$ is the pitch component of the angular momentum about the contact point, the ALIP dynamics is given by
\begin{align}
    \label{eq:ALIP}
    \begin{bmatrix}
    \dot{x} \\
    \dot{L}^y
    \end{bmatrix}
    = 
    \begin{bmatrix}
    0   & 1/(mH) \\
    mg & 0
    \end{bmatrix}
    \begin{bmatrix}
    x \\
    L^y
    \end{bmatrix},
\end{align}
where $m$ is the total mass and $H$ is the constant CoM height.
The main advantage of using the ALIP model over the LIP model is that the evolution of the angular momentum about the contact point is closer to its behavior on the full-order robot's model and the actual hardware~\cite{gong2021onestep}.

\subsubsection{Limitations}
Although ALIP does better work describing the actual behavior of the system than LIP, both are linear models subject to assumptions such as point mass body, constant CoM height, and the angular momentum about the CoM being zero during the walking gait. 

In addition, the prediction of the state at the end of the step depends on the step duration $T$. This implies that an accurate prediction would depend on the perfect timing of the touchdown event, which could only happen in ideal conditions, e.g., perfect tracking of the LL controller, point-contact foot, and non-irregular walking surfaces. 

\begin{figure}[t]
    \vspace{2mm}
    \begin{center} 
    \includegraphics[width=1.00\linewidth]{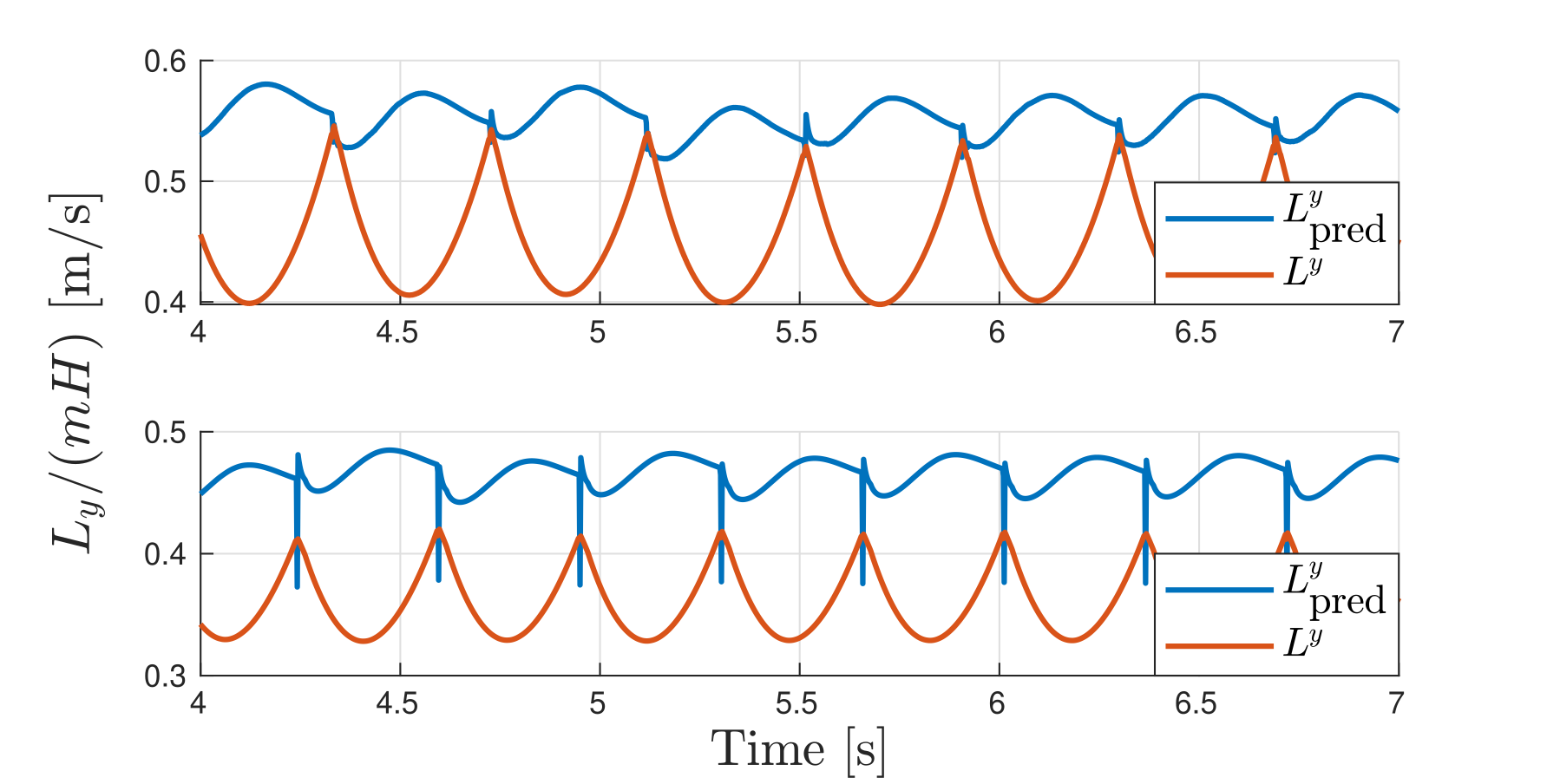}
    \end{center} 
    \vspace{-3mm}
    \caption{Prediction of $L^y$ at the end of each step under ideal (top) and non-ideal (bottom) conditions.} 
    \label{fig:limitation_ALIP}
    \vspace{-4mm}
\end{figure}

To analyze this effect, we compare the predicted value of $L^y$ scaled by $mH$ at the end of the step with the actual $L^y$ for the five-link bipedal robot Rabbit in \figref{fig:limitation_ALIP}.  We show the evolution of $L^y$ in simulation using the MuJoCo physics engine~\cite{todorov2012mujoco}. To simulate ideal conditions on the model as closely as possible, we set the geometry of the robot's links to be very thin (to emulate point contact with the ground) and use a LL feedback linearization controller with high gains to encourage better tracking and accurate touchdown timing. For the "non-ideal" conditions, we use the real geometry and dynamic properties of the robot's links (as described in~\cite{chevallereau2003rabbit}), and we use an inverse dynamics QP controller~\cite{reher2020inverse}. The results show that under non-ideal conditions, the prediction of $L^y$ at the end of the step differs significantly from its actual value. 

\subsection{Task-space LL controller}
Several approaches have been proposed in the literature to design task space controllers that consider the full-order model of the legged robot. Considering a mechanical system with configuration space $\mathcal{Q}$ and generalized coordinates $q \in \mathcal{Q}$, the equations of motion formulated using the method of Lagrange are given by:
\vspace{-1mm}
\begin{align}
    \label{eq:full-order-model}
    M(q)\ddot{q} + H(q,\dot{q}) &= Bu + J^T(q) \lambda \\
    J(q)\ddot{q} + \dot{J}(q,\dot{q})\dot{q} &= 0,  
    \label{eq:holonomic-constraints}
\end{align} 
where $D(q)$ is the inertia matrix, $H(q,\dot{q}) = C(q,\dot{q})\dot{q} + G(q) + F$ is the vector sum of the Coriolis, centripetal, gravitational, and additional non-conservative forces, B is the actuation matrix, and $J(q)$ the Jacobian of the holonomic constraints.

We denote that the system \eqref{eq:full-order-model} can be expressed in the general form \eqref{eq:hybrid_model}. Let $x=\left(q^{T}, \dot{q}^{T}\right)^{T} \in T \mathcal{Q}=\mathcal{X}$, then
\begin{align}
    \label{eq:full_order_general_form}
    &f(x)=\left[\begin{array}{c}
    \dot{q} \\
    -D^{-1}(q)\left(J^{T}(q) \lambda-H(q, \dot{q})\right)
    \end{array}\right] \\
    &g(x)=\left[\begin{array}{c}
    0 \\
    D(q)^{-1} B
    \end{array}\right].
\end{align}

The task space feedback controller tracks a set of desired trajectories of the form:
\begin{align}
    \label{eq:outputs}
    y(x)=y^{a}(x)-y^{d}(\tau(x)),
\end{align} 
where $y^a$ and $y^d$ are smooth functions, and  $y^d$ characterizes the desired behavior of the system. Upon the assumption that $y(x)$ has relative degree 1 or 2, nonlinear control methods can be applied to find a control law that drives $y(x)$ to zero, which implies the outputs converge to their target values. 

\section{Method}\label{sec:method}
This section presents the methodology for the design of the proposed learning-based hierarchical controller for bipedal locomotion. First, we introduce the overall structure of the framework. Then, we describe the learning-based HL and model-based LL components of the framework. 

\subsection{Hierarchical structure for bipedal locomotion}
The proposed learning-based framework combines the capabilities of model-based and model-free methods into a hierarchical structure to realize robust locomotion controllers for underactuated and fully-actuated bipedal robots. 
Inspired by the success of reduced-order models for the online generation of HL trajectories, we use reinforcement learning to train an HL policy that maps a reduced state space inspired by the ALIP model to a set of task space commands to generate online task space trajectories for the robot's base and end-effectors. For the LL task space controller, we use well-known model-based inverse dynamics controllers to guarantee the tracking performance of the system's outputs. 

\begin{figure}[t]
    \vspace{2mm}
    \begin{center} 
    \includegraphics[width=1\linewidth]{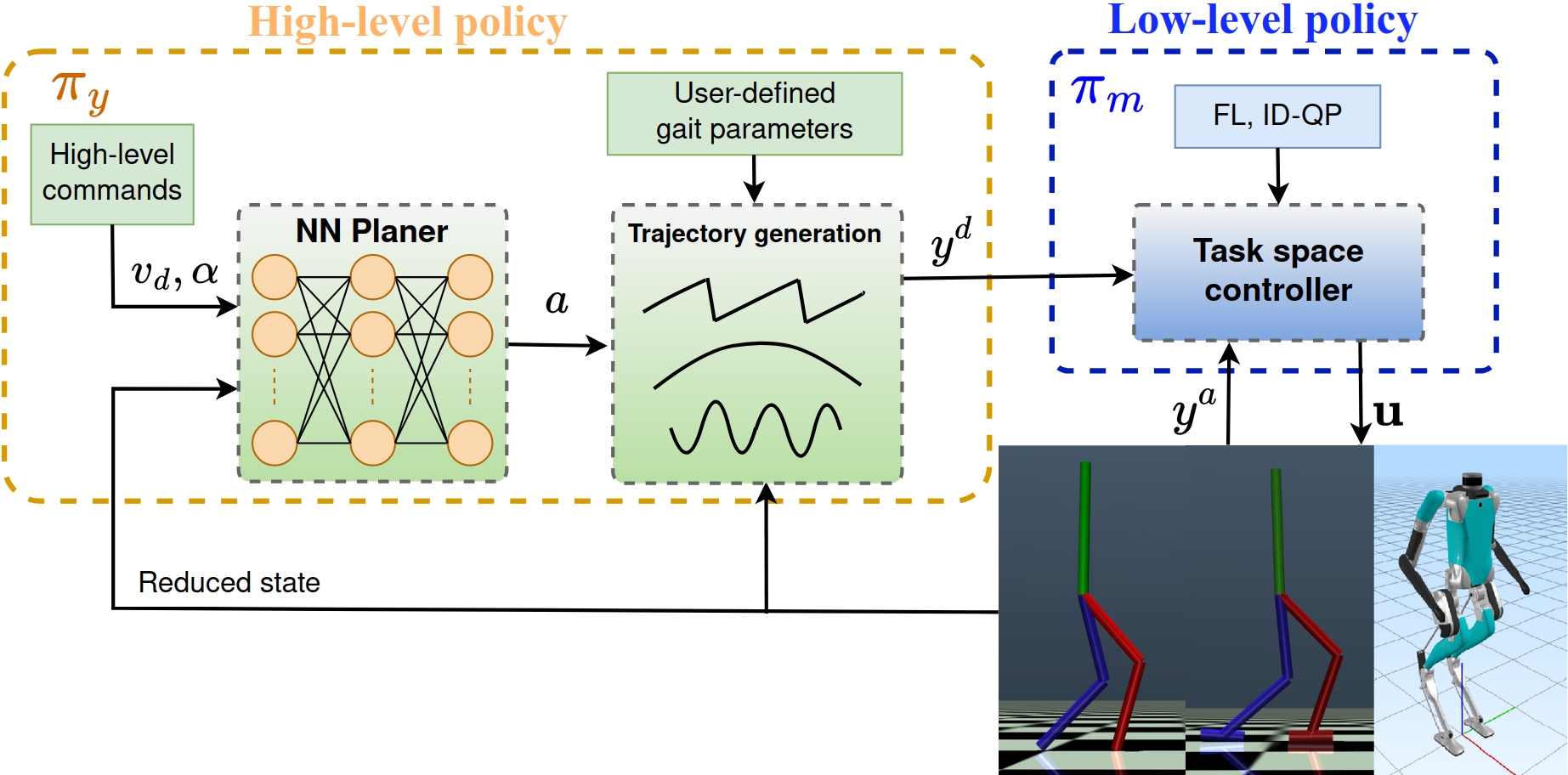}
    \end{center} 
    \caption{Overall structure of the proposed learning-based framework. The HL policy maps a reduced-order state to task space trajectories that are tracked by the LL policy.} 
    \label{fig:overall_framework}
\end{figure}

The proposed hierarchical structure is presented in ~\figref{fig:overall_framework}.  By combining the learning-based HL planner with the model-based LL controller, we obtain a robust controller capable of accurately tracking a wide range of walking speeds while preserving a good tracking performance for the task space trajectories. This significantly increases the flexibility and safety of the policy when compared with pure learning-based controllers. 

\subsection{Reinforcement Learning for High-Level Planning}
The problem of determining a motion policy for bipedal robots can be modeled as a Markov Decision Process (MDP), which consists of a tuple of components defined as
\begin{align}
    \label{eq:mdp-def}
    \mathcal{M} := (\mcs, \mca, \mathsf{P},r,\xi,\gamma).
\end{align} 
Here $\mcs$ is the state space, and $\mca$ is the set of feasible actions referred to as the action space. Specifically, at time $t$, an agent (i.e., the motion planner) takes an action $a_t\in\mca$ at state $s_t\in\mcs$, transits into the next state $s_{t+1}\in\mcs$ according to the transition probability $\mathsf{P}(s_{t+1} |s_t,a_t)$ and receives a reward $r(s_t,a_t,s_{t+1})$. Moreover, $\xi$ denotes the distribution of the initial state $s_0\in\mcs$, and $\gamma\in(0,1)$ denotes the discount factor.
The stochastic transition of the MDP process captures the random sampling of initial states in the policy training and dynamics uncertainty due to model mismatch and random interactions with the environment (e.g., early ground impacts). 

\subsubsection{Reduced-Order State Space}
Several works have already proposed using a reduced state of the robot as the observation space of the learning algorithm. However, the choice of the reduced state is made based on trial and error or empirical observations of the policy performance. In this work, we leverage recent results on the effectiveness of using angular momentum about the contact point to regulate the walking speed of biped robots~\cite{gong2021onestep}. Inspired by the ALIP model, we select the state 
\begin{align}
    s = (x, y, L^x, L^y, e_{\bar{v}^x}, e_{\bar{v}^y}, v_x^d, v_y^d, \alpha).
\end{align}
where $({x, y, L^x, L^y})$ is the ALIP state composed by the robot's base $x$ and $y$  position and the angular momentum about the contact point along the $x$ and $y$ axes, $(e_{\bar{v}_x},e_{\bar{v}_y})$ is the error between the average velocity the robot's base $(\bar{v_x}, \bar{v_y})$ and the desired robot's velocity $(v_x^d, v_y^d)$, and $\alpha$ is the terrain slope measured in radians. We denote that we use the robot's base position instead the CoM because of practical convenience for future hardware experiments. The CoM estimation on complex robots may result in noisy measurements, while the base position with respect to the contact point can be easily computed using forward kinematics. 


We assume the slope of the terrain is known by the learning agent. This assumption is reasonable since most of the bipedal robots available for research and commercial applications are equipped with perception systems to map the surrounding environment. Even in the absence of perception systems, proprioceptive approaches could be used to accurately estimate the terrain slope based on the orientation of the robot's base and feet, as we have shown in simulation and hardware in our previous work~\cite{castillo2022reinforcement}.

\subsubsection{Task Action Space}
The action $a \in \mathcal{A}$ is chosen to be 
\begin{align}
    a = (p^{x}_{\text{sw},T}, p^{y}_{\text{sw},T}, q_\phi, 
    h^d) 
\end{align}
where $p^{x}_{\text{sw},T}, p^{y}_{\text{sw},T}$ correspond to the position of the swing foot w.r.t. the robot's base at the end of the swing phase $T$, i.e., the landing position of the swing foot, $q_\phi$ is the absolute torso pitch angle, and $h^d$ is an offset added to the nominal height of the robot's base w.r.t. the stance foot. This selection of the action space encourages the flexibility of the policy to exploit the natural nonlinear dynamics of the biped robot and enhance the robustness of the policy under big disturbances, sudden speed changes, and walking at high speeds, as it will be shown in section \ref{sec:results}. 

The HL actions $a$ are used to generate smooth task space trajectories for the robot's floating base and end-effectors. 
The trajectory for $p^{x,y}_{\text{sw}}$ is generated using a minimum jerk trajectory of a straight line segment connecting $p^{x,y}_{\text{sw},0}$ with $p^{x,y}_{\text{sw},T}$. The swing foot position at the beginning of the walking step, $p^{x,y}_{\text{sw},0}$ is computed using Forward Kinematics (FK) and updated at every touch-down event. The desired position for the swing foot at landing, $p^{x,y}_{\text{sw},T}$ is updated by the HL policy at the frequency of $30 Hz$. The vertical trajectory of the swing foot position w.r.t. the robot's base is generated using a 5th order Bézier Polynomial parameterized by the vertical position of the foot at the beginning $(p^z_{\text{sw},0})$ and the end $(p^z_{\text{sw},T})$ of the step, and the high foot clearance $p^z_{sw,T/2}$. For flat ground terrain, we have
\begin{align}
    p^z_{\text{sw},T} = -h^d
\end{align}
We update $p^z_{\text{sw},T}$ by 
\begin{align}
    p^z_{\text{sw},T} = -h^d + p^x_{\text{sw},T} * \tan(\alpha) - p^z_\text{off},
\end{align}
where $\alpha$ is the terrain slope and $p^z_\text{off} = 0.005$m is a small offset added to guarantee the swing foot makes contact with the ground. 


The Neural Network chosen to parameterize the HL policy is a Recurrent Neural Network with 2 hidden layers, each layer with 128 units for the case of 2D robots and 256 units for 3D robots. The hidden layers use the ReLU activation function, and the output layer is bounded by the sigmoid activation function and a scaling factor to constrain the maximum value of the HL commands. 

\subsection{Low-level task space controller} \label{subsec:low-level_control}
The LL task space controller is designed using standard techniques of the nonlinear systems control literature. In particular, we implement two types of model-based controllers i) Feedback Linearization (FL), and ii) Inverse Dynamics with QP formulation (ID-QP).
We evaluate the performance of the HL policy with different LL controllers and show that the learned policy is robust to any choice of the LL controller. The purpose of this evaluation is to demonstrate the versatility of the task space-based HL planner to adapt to different LL control structures without affecting the performance of the learned policy. This also provides more flexibility for the designer to use any LL control approach at their convenience. For instance, FL is easy to implement and requires less computation time, but it is known to be hard to implement on real hardware. Therefore, FL could be used during the training process of the HL policy, while any suitable ID-QP formulation could be used for hardware experiments.

For more details, we refer the reader to~\cite{reher2020inverse, delprete2015prioritized}, where several QP formulations for bipedal locomotion are proposed with successful applications to real hardware. In this work, we use the most basic case of the ID-QP formulation in ~\cite{reher2020inverse} for the 2D robots and the Task Space Inverse Dynamics (TSID) formulation in ~\cite{reher2020inverse} for the 3D robot Digit. 

\subsection{Learning procedure} \label{learning_procedure}
The reinforcement learning algorithm we use in this work is an implementation of the Proximal Policy Optimization~\cite{schulman2017proximal} algorithm with parallel experience collection, input normalization, and fixed covariance. The algorithm shares the same code base as the implementations in~\cite{green2021learning} and \cite{siekmann2020learning}. 

 For each episode, the initial state of the robot is set randomly from a normal distribution about an initial pose corresponding to the robot standing in the double support phase. One iteration of the HL policy corresponds to the interaction of the learning agent with the environment. The HL policy takes the reduced-order state $s \in \mathcal{S}$ and computes an action $a \in \mathcal{A}$ that is converted in desired task space trajectories $y^d$ at the time $t$. The reference trajectories are then sent to the LL task space controller. The LL control loop runs at a frequency of 1 KHz, while the HL planner runs at 33 Hz. The maximum length of each episode is 300 steps, which corresponds to 9 seconds of simulated time.  

The episode has an early termination if any of the following conditions are violated: 
\begin{align}
    &|q_\phi| < 1 \text{rad}, \quad h < 0.5 \text{m}.
\end{align}

The simple reward function \eqref{eq:reward} adopted in this work is designed to keep track of the target walking speed while realizing a stable walking gait. In particular, the terms $r_{v_x}, r_{v_y}$ encourage the tracking of the longitudinal and lateral target speeds. Since the torso pitch angle is part of the learning action space, the term $r_{L_\textrm{CoM}}$ encourages the policy to avoid excessive changes in the torso orientation without explicitly restricting the torso pitch angle. Finally, the term $r_{a}$ encourages the policy to avoid excessive variations between the last action and the current action. This avoids unnecessary overshooting in the commanded actions that may produce risky behaviors during the walking gait. The weighted reward function is given as:
\begin{align}
    \label{eq:reward}
    \mathbf{r} = \mathbf{w}^T [ r_{v_x}, r_{v_y}, r_{L_\textrm{CoM}}, r_{a}]^T,
\end{align}
where 
\begin{align}
    \label{eq:reward_detailed}
    r_{v_x} &= \exp{({-\norm{\bar{v}_x - v_x^d}}^2)}\\
    r_{v_y} &= \exp{({-\norm{\bar{v}_y - v_y^d}}^2)}\\
    r_{L_\textrm{CoM}} &= \exp{({-\norm{L_\textrm{CoM}}}^2)}\\
    r_{a} &= \exp{({-\norm{a_k - a_{k-1}}}^2)}.
\end{align}
and $\mathbf{w}^T$ is a vector of weights corresponding to each reward term. 
For 2D robots we use $\mathbf{w}^T=[0.6, 0, 0.2, 0.2]$ while for 3D robots we use $[0.3, 0.3, 0.2, 0.2]$.

\section{Illustration example}\label{sec:illustration}
In this section, we show the proposed method can be generalized to both underactuated and fully actuated robots without any changes to the structure of the HL planner policy. Moreover, we demonstrate the framework can be applied in 2D and 3D bipedal robots. We use 3 different robots.

\textbf{Rabbit} is a five-link, planar underactuated bipedal robot with point feet and four actuated joints, two in the hips and two in the knees. Despite its simple mechanical structure, Rabbit still provides a suitable representation of biped locomotion, which is the reason it has been considered as a test bed for advanced control theory in the field of legged robots~\cite{chevallereau2003rabbit}.

\textbf{Walker2D} is a seven-link, planar, fully actuated bipedal robot with 6 actuated joints, two in the hips, two in the knees, and two in the ankles. The additional degrees of freedom at the ankles enable the robot to realize human-like walking gaits and balancing.

\begin{figure}[h]
    \begin{center} 
    \includegraphics[width=1\linewidth]{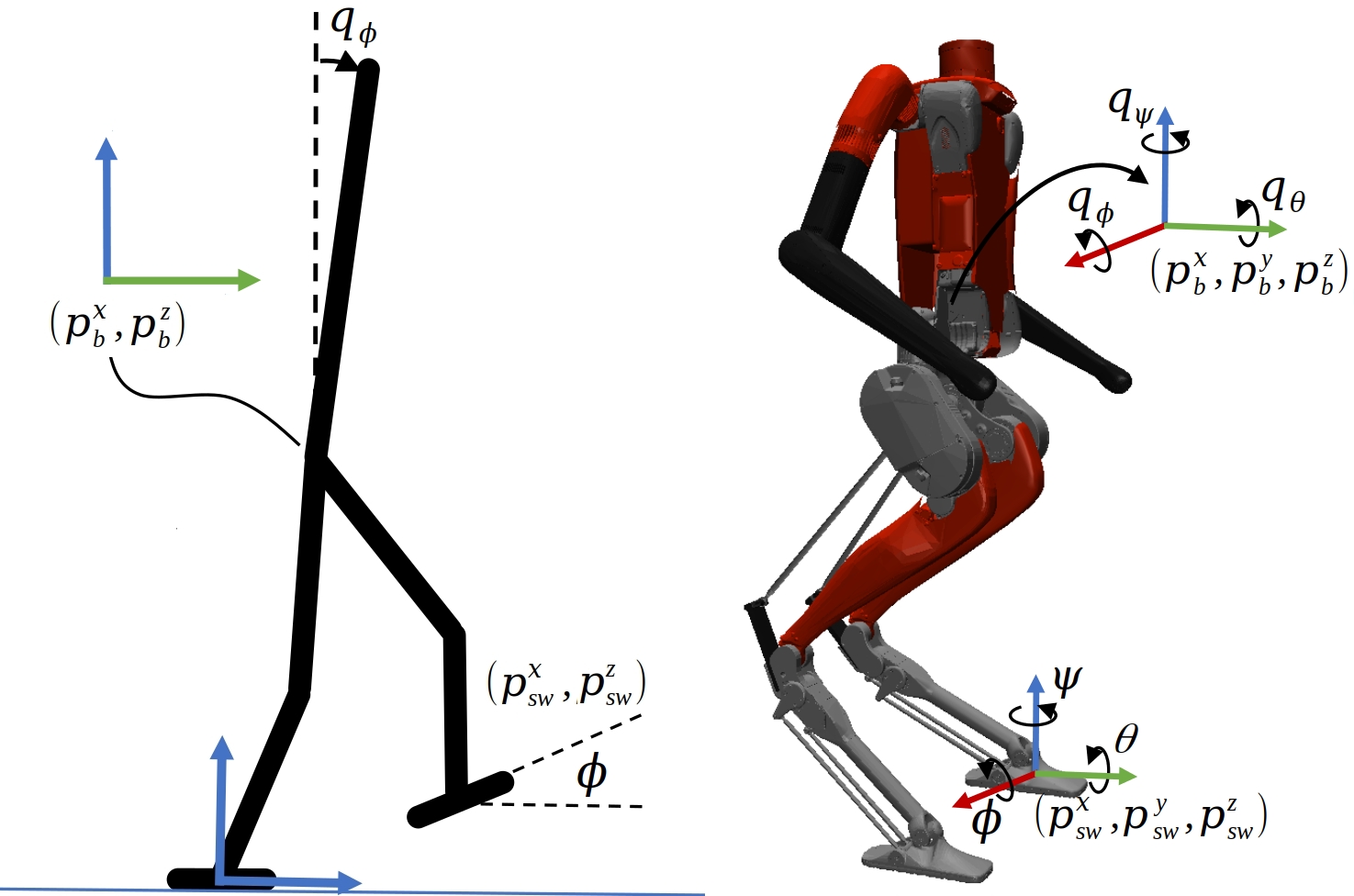}
    \end{center} 
    \caption{Schematics of the robots Walker2D and Digit.} 
    \label{fig:rabbit_schematic}
\end{figure}

Schematics of the Walker2D and Digit
are shown in \figref{fig:rabbit_schematic}. Rabbit shares the same design and structure as Walker2D without the feet and ankle joints.

\textbf{Digit} is a 3D fully actuated bipedal robot with 30 DoF and 20 actuated joints. Each leg has six actuated joints corresponding to the motors located on the robot’s hip, knee, and ankle and two passive joints corresponding to the robot’s shin and tarsus joints. In addition, it has four actuated joints per arm corresponding to the shoulder and elbow joints. \figref{fig:rabbit_schematic} shows the kinematic structure of Digit.

\subsection{Task-space outputs for the LL controller.}
The set of task space outputs of relative degree 2 described by equation \eqref{eq:outputs} to characterize the walking gait of the biped robot are defined as follows:
\begin{align}
    y_{2}^{a}(q):=\left[\begin{array}{c}
    q_\phi \\
    h \\
    p_{sw}^x\\
    p_{sw}^y\\
    p_{sw}^z\\
    \phi_{sw}
    \end{array}\right] \rightarrow
    \left(\begin{array}{c}
    \text { torso pitch angle } \\
    \text { base height } \\
    \text { swing foot $x$ } \\
    \text { swing foot $y$ } \\
    \text { swing foot $z$ } \\
    \text { swing foot pitch }
    \end{array}\right)
\end{align}

This selection of outputs is common in the field of bipedal locomotion. The first five outputs ($q_\phi, h, p_{sw}^x, p_{sw}^y, p_{sw}^z$) are valid for both underactuated and fully actuated robots. 
However, for an underactuated robot, it is not possible to control the horizontal position of the robot's base or CoM. Therefore, the evolution of the base velocity is indirectly controlled by the HL learned policy through the  planning of touchdown position. 
In the case of fully actuated robots, we also consider the sixth output to control the swing foot pitch angle to be  parallel to the walking surface. This contributes to reducing disturbances at the touchdown event. Although we could add an additional output ($\phi_{sw}$) to control the horizontal position and velocity of the robot's base, we prefer to rely on the HL planning to control the robot's speed indirectly. The objectives of this choice of design are twofold:
i) devise a general framework for both underactuated and fully-actuated robots that share the same structure for the HL policy independently of the particular design of the bipedal robot.
ii) simplify the design of the controller and avoid limitations of the torque ankle to control the robot's base position. Although this effect could not be significant for quasi-static locomotion gaits, it does matter when realizing agile and dynamic locomotion.

\section{Simulation results}\label{sec:results}
In this section, we show the performance of the learned HL policy under different testing scenarios with three different robot models, including Rabbit, Walker2D, and Digit. Moreover, we analyze the contribution of the HL policy to the robustness of the walking gait, and we compare our method with similar model-based and model-free approaches.

\subsection{Speed tracking for different velocity profiles.} \label{subsec:results_speed_tracking}
\begin{figure}[h]
    \begin{center} 
    \includegraphics[width=1.0\linewidth]{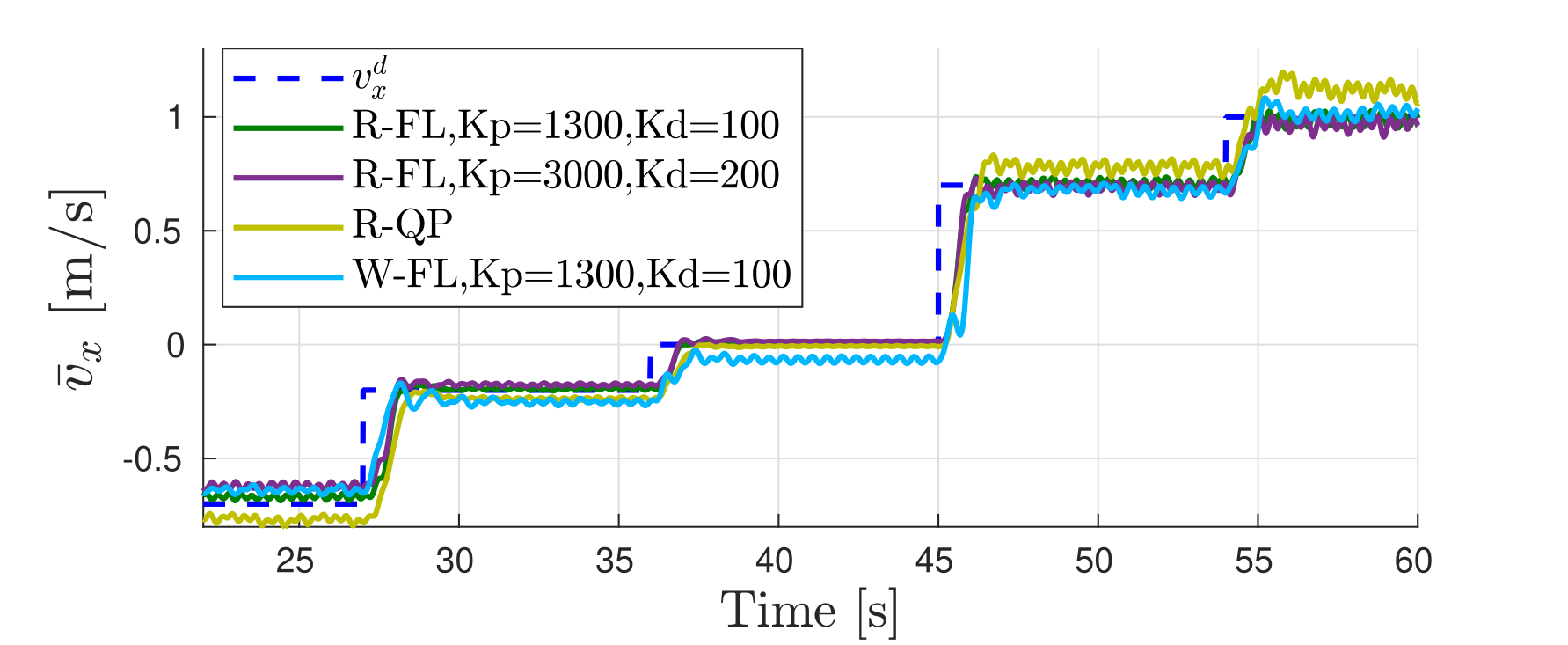}
    \end{center} 
    \vspace{-4mm}
    \caption{Velocity tracking performance of the learned policy with different LL controllers. The prefix R/W is used to differentiate a policy for Rabbit or Walker2D.} 
    \vspace{-4mm}
    \label{fig:speed_tracking}
\end{figure}

We test the learned policy for tracking a velocity profile in different directions. \figref{fig:speed_tracking} shows the velocity tracking performance of the learned HL policy for the robots Rabbit and Walker2D. To evaluate the robustness of the policy under different LL controllers, we test the same policy with the ID-QP controller and the FL controller with different tracking gains. The results show the policy effectively tracks the walking speeds in the range $[-1,1]$ m/s, even with aggressive changes in the velocity profile.

\begin{figure}[h]
    \begin{center} 
    \includegraphics[width=1\linewidth]{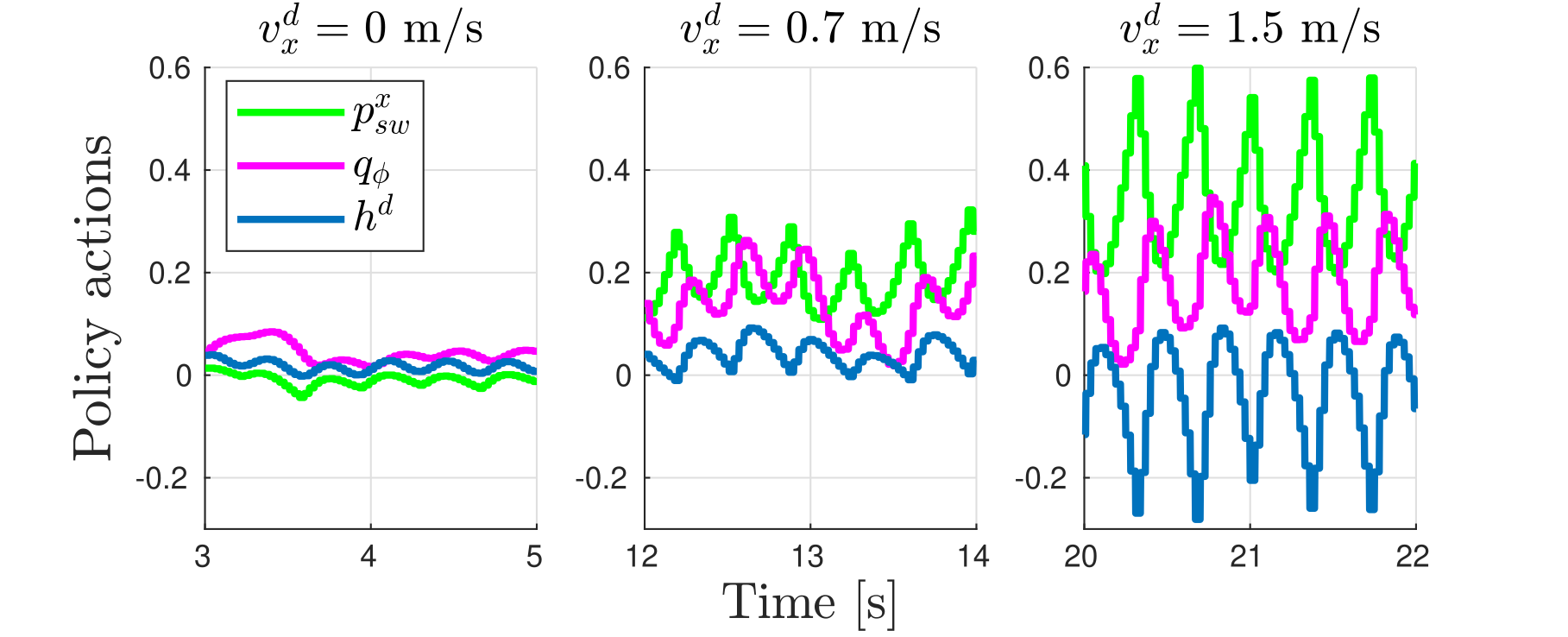}
    \end{center} 
    \vspace{-4mm}
    \caption{Contribution of the policy actions for different speeds.} 
    \vspace{-4mm}
    \label{fig:policy_actions}
\end{figure}

To highlight the contribution of the choice of action space, we present in \figref{fig:policy_actions} the actions computed by the HL policy for different commanded velocities. When a steep change in the desired velocity is commanded, the policy uses the torso orientation to compensate for variations in the robot's speed and angular momentum. We denote that these behaviors are not enforced during the training process but arise naturally from the insightful design of the proposed framework. Interestingly, some of these strategies are also observed in human locomotion~\cite{bennett2010angular}.

\subsection{Comparison with ALIP model-based approach}
\begin{figure}[h]
    \vspace{-4mm}
    \begin{center} 
    \includegraphics[width=1\linewidth]{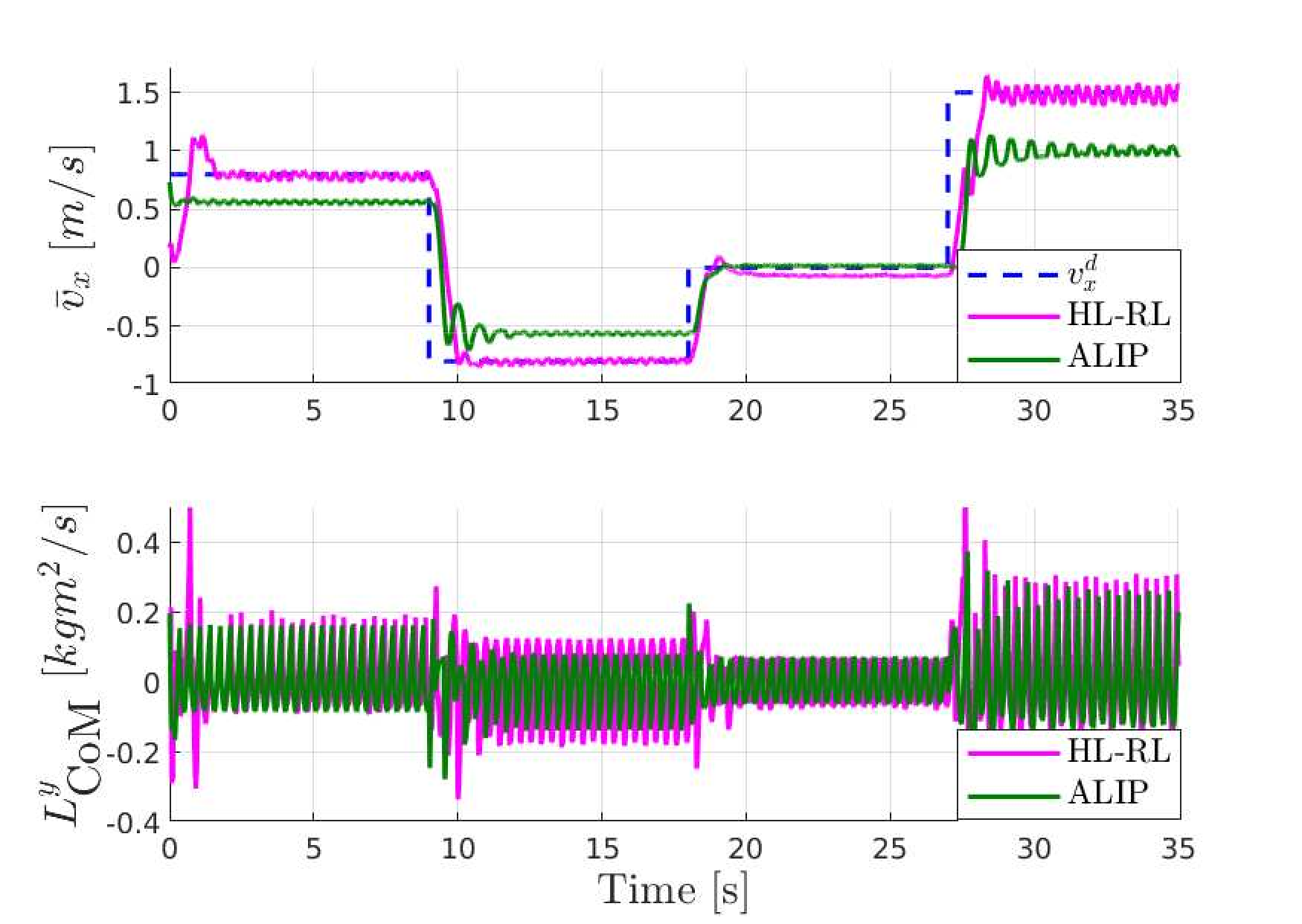}
    \end{center} 
    \vspace{-4mm}
    \caption{Comparison of HL-RL with ALIP-based controller.} 
    \vspace{-4mm}
    \label{fig:comparison}
\end{figure}

To assess the advantages of the proposed HL planner with respect to pure model-based approaches, we compare the performance between our learning-based controller and the ALIP-based controller. \figref{fig:comparison} (top) shows the HL-RL policy (ours) outperforms the model-based controller in tracking a velocity profile, especially for high speeds. We also show the variation of $L_{\textrm{CoM}}$ (bottom) to denote the trade-off the policy learns between minimizing $L_{\textrm{CoM}}$ and tracking $v_x^d$. For small speeds, $L_{\textrm{CoM}}$ looks quite similar for both controllers. For high speeds, the policy learns to prioritize speed tracking over minimizing $L_{\textrm{CoM}}$. This behavior is encouraged by the selection of weights in the reward function \eqref{eq:reward}.

\subsection{Comparison with other RL-based approaches}
To highlight the contribution of the proposed framework in terms of speed tracking for 3D bipedal robots and sample efficiency, we compare our method with our previous work in \cite{castillo2022reinforcement} and the end-to-end learning approach presented in \cite{siekmann2020learning}. Although there are no end-to-end learning approaches implemented on the Digit robot in the literature, there are several works that have done so with the robot Cassie, which shares the same leg morphology as Digit. Therefore, we choose to implement the method in \cite{siekmann2020learning} with Digit as it focuses on speed tracking, which makes it more comparable to our method. Moreover, the framework in \cite{siekmann2020learning} is the base for several SOTA end-to-end learning approaches for bipedal locomotion \cite{siekmann2021sim, duan2021learning, duan2022sim-to-real}. Finally, the code implementation for \cite{siekmann2020learning} is publicly available online, which makes the comparison as fair as possible in terms of the reproducibility of their work. 

\begin{figure}[h]
    \vspace{-4mm}
    \begin{center} 
    \includegraphics[width=1.0\linewidth]{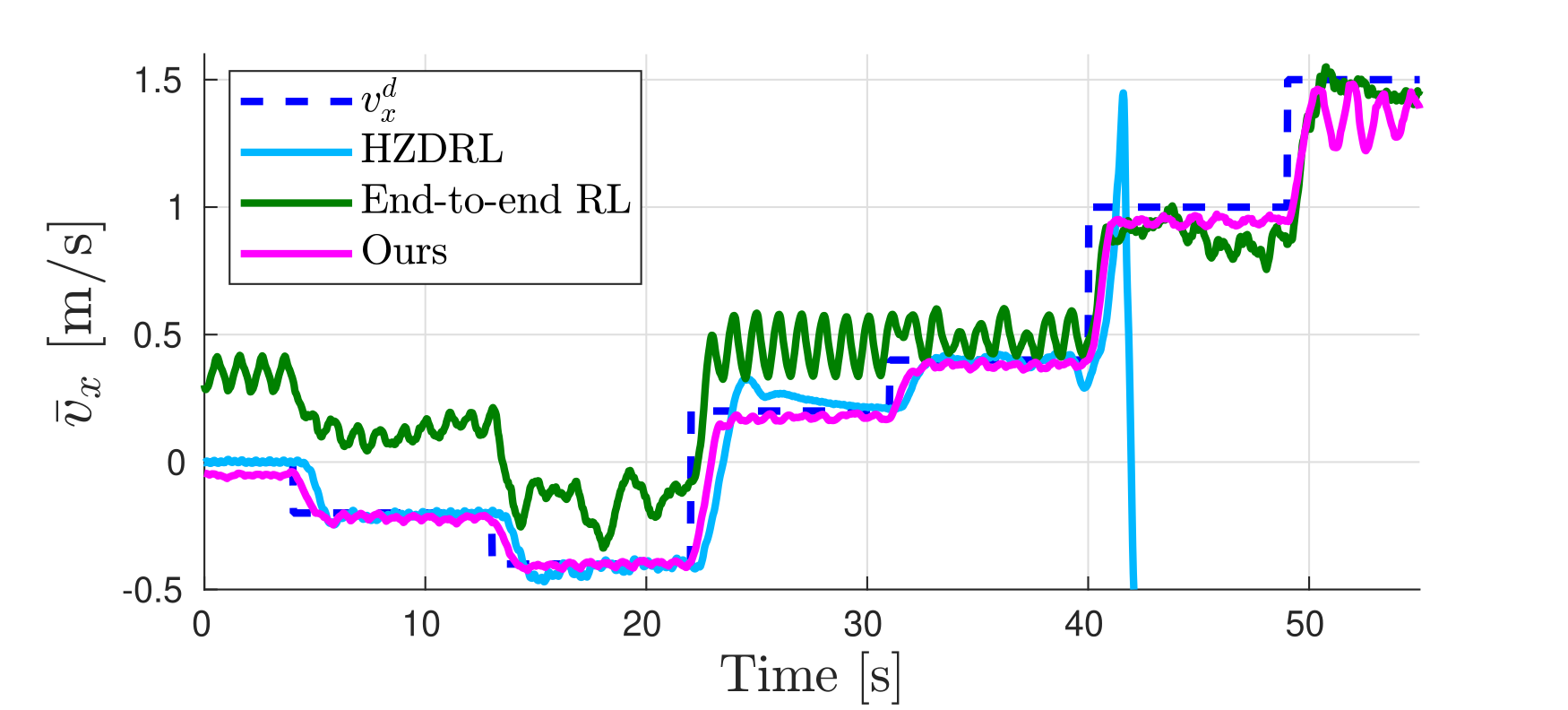}
    \end{center} 
    \vspace{-4mm}
    \caption{Comparison of speed tracking performance with SOTA RL-based approaches.} 
    \vspace{-4mm}
    \label{fig:comparison_vx}
\end{figure}

In \figref{fig:comparison_vx}, we present the comparison results for speed tracking on flat ground with the robot Digit using our learned HL policy, the HZDRL controller in \cite{castillo2022reinforcement}, and the end-to-end RL policy in \cite{siekmann2020learning}.  We observe that the HZDRL controller fails, i.e., the robot falls, for speeds higher than 0.5 m/s, while the end-to-end RL controller fails to track low speeds accurately and realizes a non-smooth walking motion that causes higher variance in the speed profile. This effect may be caused by the reference trajectory used to guide the learning. In our implementation of \cite{siekmann2020learning}, we use a reference trajectory corresponding to Digit walking forward at a speed of 0.8 m/s. For more details, the reader can refer to \cite{siekmann2020learning}. Our learned policy can successfully track the desired walking speed in a significantly wider range compared with the other methods. Additional testing with Digit demonstrates our controller can handle desired walking speeds in the range $v_x \in [-1.0, 1.5]$m/s and $v_y \in [-0.5, 0.5]$m/s, including combinations of both, i.e., diagonal walking, as can be seen in the accompanying video submission. 

\begin{figure}[h]
    \vspace{-3mm}
    \begin{center} 
    \includegraphics[width=1\linewidth]{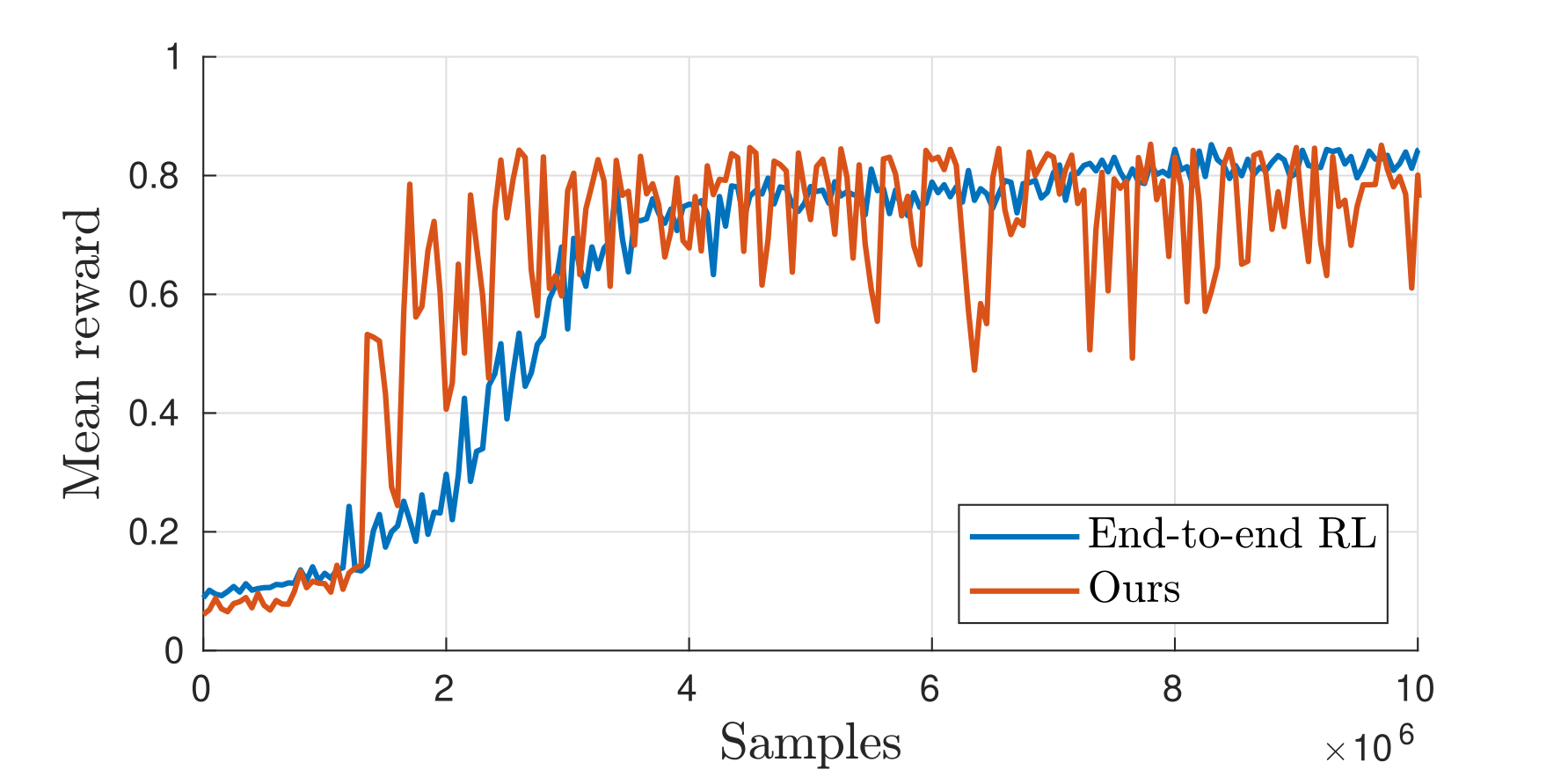}
    \end{center} 
    \vspace{-4mm}
    \caption{Comparison of sample efficiency between traditional end-to-end RL and our proposed method. The higher variance in the reward is caused by the effect of the exploration noise in the task-space actions, which allows the policy to explore a more diverse set of behaviors during training. } 
    \vspace{-4mm}
    \label{fig:comparison_reward}
\end{figure}

In terms of sample efficiency, \figref{fig:comparison_reward} shows our approach uses fewer data samples to successfully train a robust policy when compared with the end-to-end learning approach. Although this is not a surprising result as several authors have studied the effects of task space learning in the data sample efficiency \cite{duan2021learning, bellegarda2019training}, to the best of our knowledge, our method is the only task space approach that learns to walk from scratch and has been successfully implemented in 3D bipedal robots without the need of previously computed reference trajectories, e.g., the gait library used in \cite{duan2021learning}.

\subsection{Robustness on challenging terrain}
\begin{figure}[h]
    \begin{center}
    \vspace{-1mm}
    \includegraphics[width=1\linewidth]{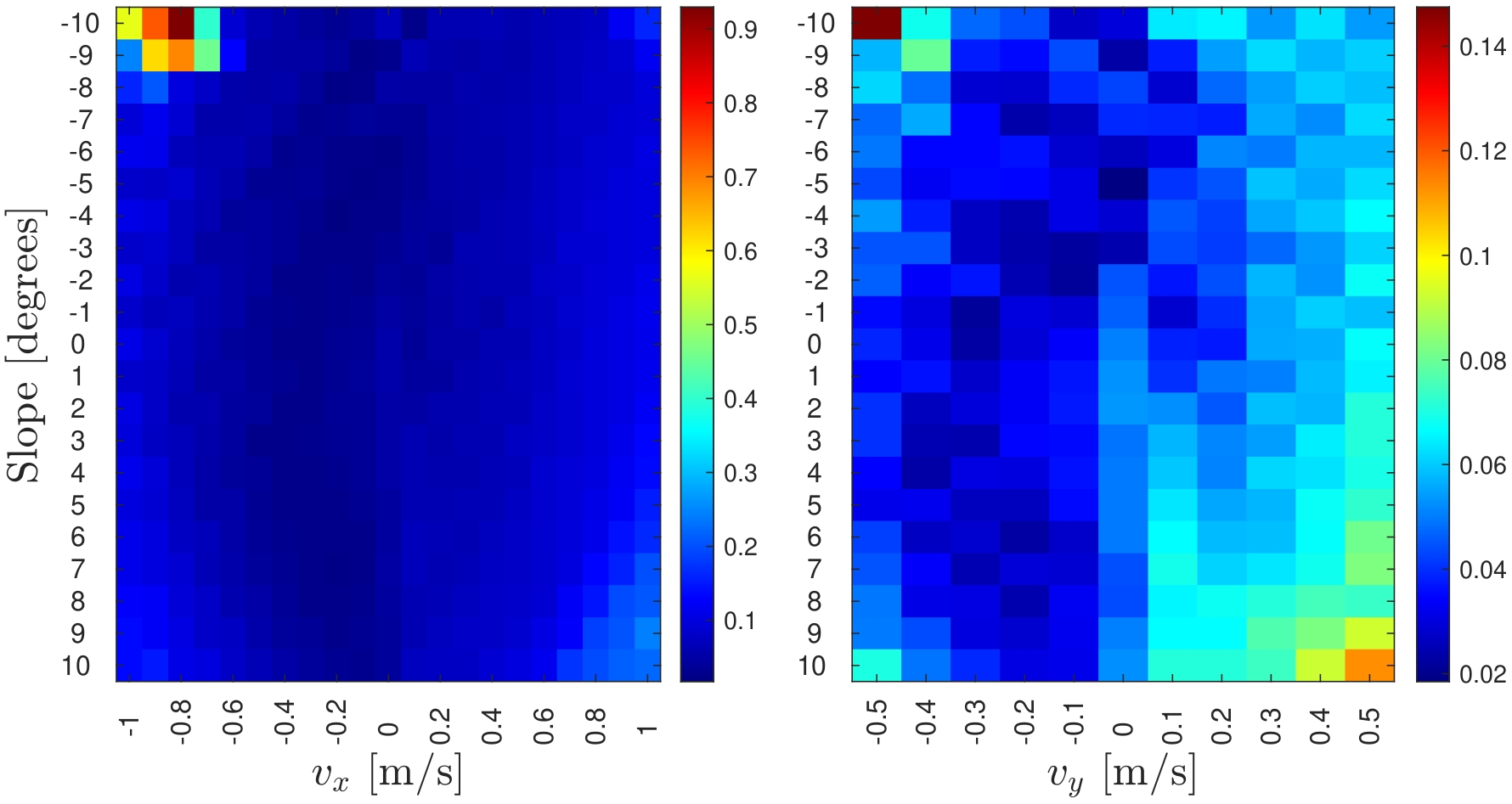}
    \end{center} 
    \vspace{-4mm}
    \caption{Robustness of the learned policy under different conditions of terrain slope and target speed.} 
    \vspace{-1mm}
    \label{fig:slopes_digit}
\end{figure}

We test our controller in terrains with slopes up to 20 degrees for Rabbit and Walker2D and 10 degrees for Digit. \figref{fig:slopes_digit} shows a grid plot with the Root Mean Squared Error between 
$\bar{v}_x$ and $v_x^d$ for  Digit with $v_x^d \in [-1.0,1.0]$m/s, $v_y^d \in [-0.5, 0.5]$m/s, and $\alpha \in [-10,10]$ degrees.
We denote that during training, we only use $\alpha \in [0,10]$. Yet, we test the policy in an extended range of slopes to highlight the robustness and interpolation capability of the policy to scenarios not seen during training. 
In this scenario, we define the robustness of the policy by its capability to keep track of $v_x^d, v_y^d$ under challenging conditions. The results show that the learned policy keeps good tracking of $v_x^d$ in almost all conditions, except for combinations where the robot is walking backward at high speed on very steep terrain. 

A higher error in the tracking of $v_y^d$ is caused by the higher variance of the instantaneous velocity along the $y$ axis, which is expected in bipedal locomotion as the robot's torso tends to oscillate more about the sagittal plane. We also notice there is a higher error when tracking positive speeds along the $y$ axis. This effect may be caused by a biased created during training by the exposure of the policy to more samples with negative commands of $v_y$.


\subsection{Robustness against disturbances.}
\begin{figure}[h]
\vspace{-4mm}
    \begin{center} 
    \includegraphics[width=1\linewidth]{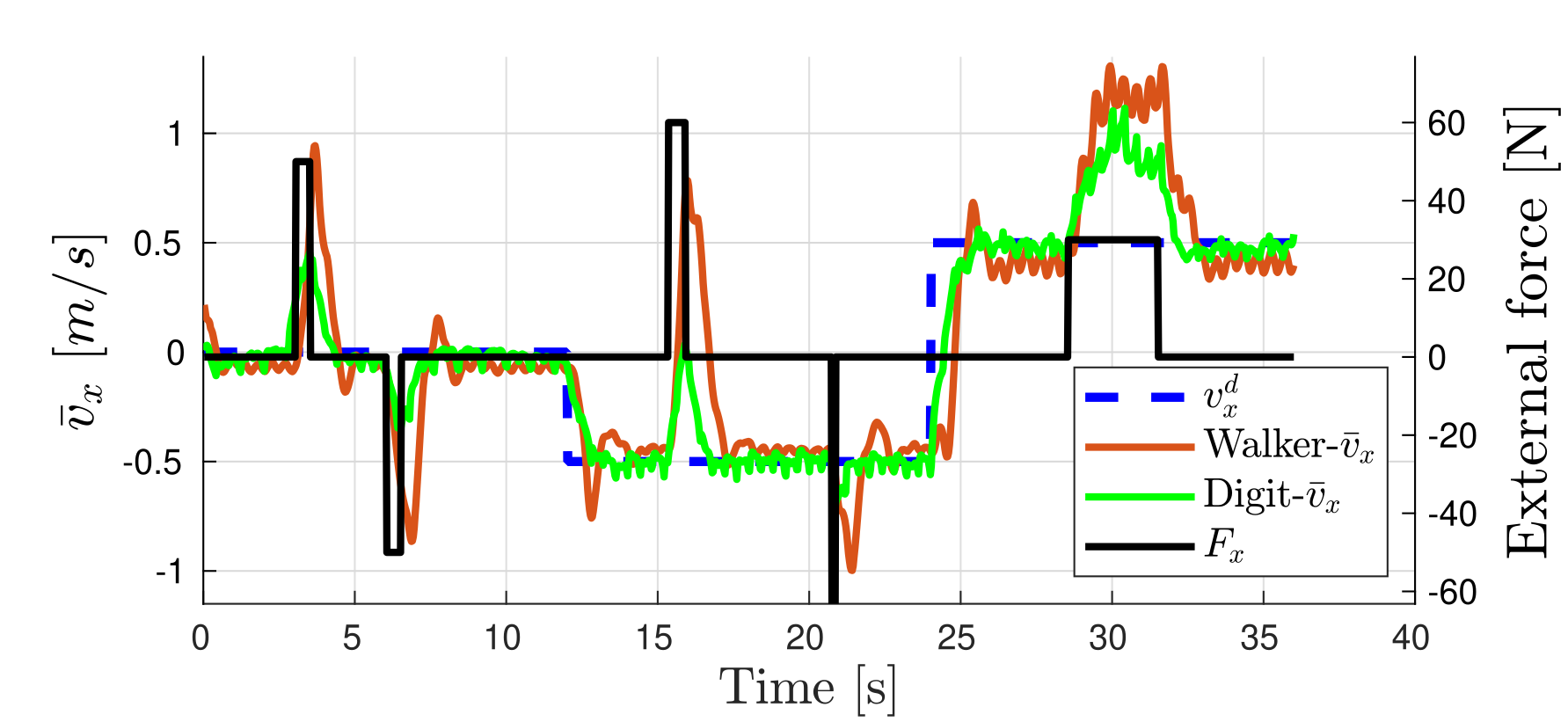}
    \end{center} 
    \vspace{-4mm}
    \caption{Robustness test to external disturbances.} 
    \vspace{-4mm}
    \label{fig:robustness_force}
\end{figure}

The trained policy is also subjected to extensive tests against external disturbances $F_x \in [-80,60] $N applied in both forward and backward directions with duration $t \in [0.15,3]$s. \figref{fig:robustness_force} shows the policy reacts effectively to all the applied disturbances without falling and maintaining the tracking of the desired walking speeds. In some scenarios, the policy uses interesting combinations of the task space outputs to realize effective strategies to reject the disturbances, e.g., bending forward/backward to absorb the impact. The results are similar for Walker2D and Digit. More details can be seen in the accompanying video submission. 

\section{Conclusion}\label{sec:conclusion}
In this work, we present a simple and effective learning-based hierarchical approach to realize robust locomotion controllers for bipedal robots.
The design of the HL policy is inspired by the reduced-order state of the ALIP model, and a set of task space commands that include the step length, torso orientation, and height. This insightful choice of state and action spaces results in a compact policy that learns effective strategies for robust and dynamic locomotion. We show the HL planner is agnostic to the choice of LL controller, and its application is general to underactuated and fully actuated 2D and 3D robots. 
Finally, we show the learned policy tracks a wide range of speeds even under challenging conditions, such as external forces and slopes up to 20 degrees.
Future work will focus on hardware experiments with the robot Digit and the extension of the proposed hierarchical framework to integrate different behaviors such as balancing, climbing stairs, and walking over stepping stones.



\bibliographystyle{IEEEtran}
\bibliography{references.bib}  

\end{document}